\documentclass[runningheads]{llncs}

\usepackage{eccv}

\usepackage{eccvabbrv}
\usepackage{multirow}   
\usepackage{graphicx}
\usepackage{booktabs}

\usepackage[accsupp]{axessibility}  

\usepackage[pagebackref,breaklinks,colorlinks,citecolor=eccvblue]{hyperref}
\usepackage{hyperref}

\usepackage{orcidlink}

\newcommand\ours{\texttt{BEACON}\xspace}

\begin{document}

\title{\ours: Behavior and Appearance Control for Subject-Specific Video Generation} 

\titlerunning{Abbreviated paper title}

\author{
Baptiste Pokrzywa\inst{1,2},
Nabyl Quignon\inst{3},
Yara Bahram\inst{1},
Muhammad Osama Zeeshan\inst{1},
Antitza Dantcheva\inst{3}\orcidlink{0000-0003-0107-7029},
Eric Granger\inst{1}\orcidlink{0000-0001-6116-7945}
}

\institute{
LIVIA, École de technologie supérieure, Montreal, Canada
\and
CentraleSupélec, Université Paris-Saclay, Gif-sur-Yvette, France
\and
Inria, Université Côte d'Azur, Sophia Antipolis, France
}

\authorrunning{B. Pokrzywa et al.}

\institute{
LIVIA, École de technologie supérieure, Montreal, Canada
\and
CentraleSupélec, Université Paris-Saclay, Gif-sur-Yvette, France
\and
Inria, Université Côte d'Azur, Sophia Antipolis, France
}

\maketitle

\begin{abstract}
Generating human-centric videos that preserve both visual identity and person-specific expressive behavior remains a fundamental challenge. In addition to reproducing appearance, a model must replicate the facial behaviors that characterize how a subject expresses emotion over time. However, most state-of-the-art methods condition generation on a single reference image, which contains no information about these temporal dynamics. As a result, they tend to preserve the subject’s visual identity but often produce expressions with limited variation and weak subject specificity. 
To mitigate this issue, we introduce \ours, a lightweight framework for person-specific video generation that produces more expressive videos by disentangling visual identity from expressive behavior. \ours conditions generation on two complementary signals: a reference image encoding the identity and a reference video capturing subject-specific facial dynamics. By conditioning on these complementary signals, \ours generates videos that better preserve both the subject’s appearance and characteristic facial dynamics, while also supporting identity-expression transfer. 
Our experiments\footnote{Our code is included in supplementary materials and will be made public.} on the MEAD and RAVDESS datasets show that by fine-tuning on approximately 2,000 pairs and updating about 1\% of the pretrained Wan video diffusion model, \ours improves facial expressivity over state-of-the-art video generation methods while maintaining competitive identity preservation.


\end{abstract}

\section{Introduction}

\begin{figure}[!h]
    \centering
    \includegraphics[width=0.9\linewidth]{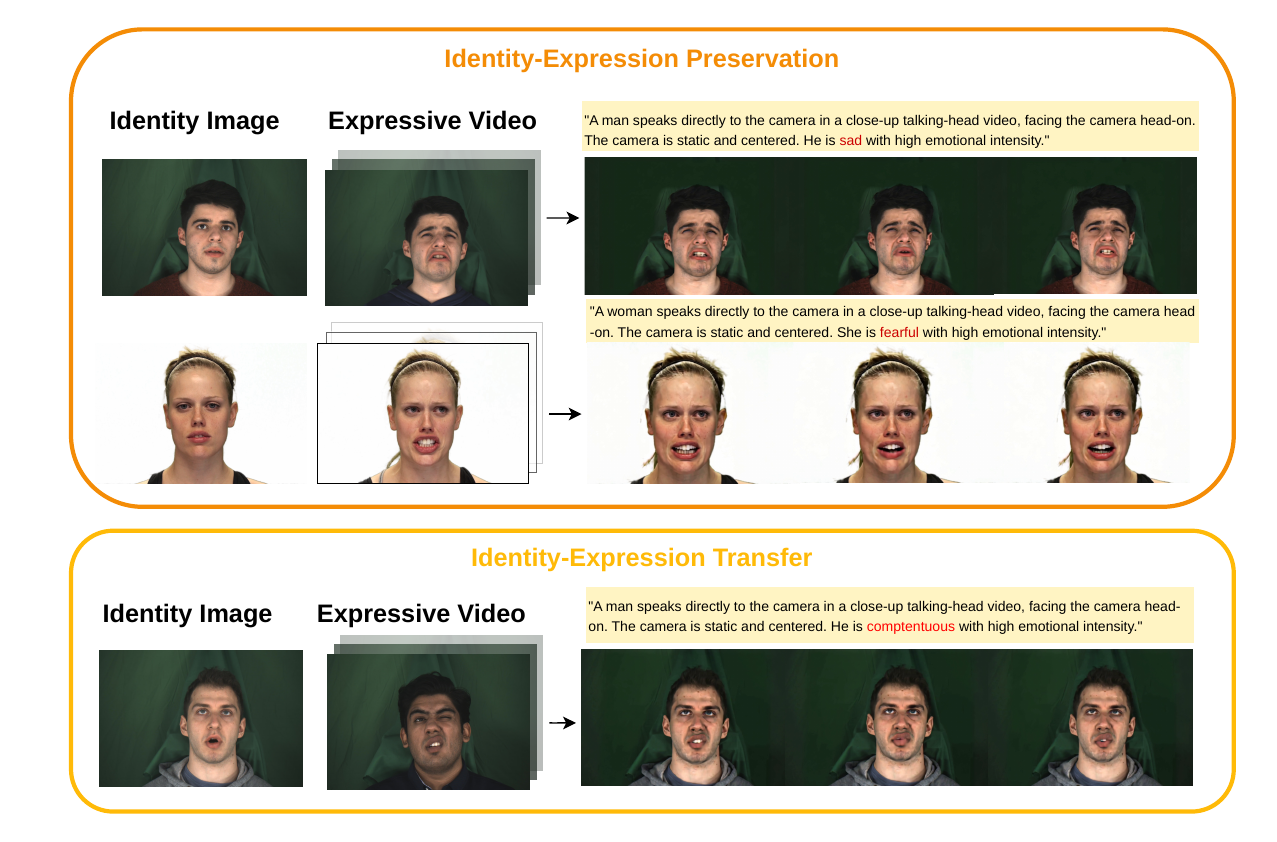}
    \caption{Given a reference image encoding visual identity, a reference video capturing subject-specific facial dynamics, and a text prompt, \ours generates expressive videos that preserve both the subject's appearance and characteristic facial behaviour (top), while enabling identity-expression transfer across different subjects (bottom).}
    \label{fig:main}
\end{figure}

Large-scale pre-trained video diffusion models \cite{wan2025wanopenadvancedlargescale, lin2024opensoraplanopensourcelarge, hong2022cogvideolargescalepretrainingtexttovideo} have enabled many controllable video generation tasks, including identity-preserving video generation. The goal is to generate realistic videos of a target individual while faithfully preserving the person's visual appearance from a reference signal. Such capabilities have numerous applications in virtual avatars, content creation, and human-computer interaction.
Despite recent advances, preserving a person's visual identity is only one aspect of human-centric video generation. State-of-the-art methods \cite{he2024idanimator, xue2025stand, lai2026slot, liu2025phantomsubjectconsistentvideogeneration} primarily focus on controlling who appears in the generated video, while providing limited control over how that person behaves and expresses spontaneous emotions over time. Affective behaviour is dynamic and spatio-temporal in nature. In a video, emotions may be conveyed through facial and vocal expressions, gaze, head motion, gestures, and their temporal evolution \cite{11455189, pantic2005dynamics, kaisiyuan2020mead}. For example, two individuals may smile or express pain with substantial differences in timing, intensity, asymmetry, or motion patterns.

SoTA methods for human-centric video generation\cite{xue2025stand, liu2025phantomsubjectconsistentvideogeneration, lai2026slot, yuan2025cons} tend to preserve the subject’s visual identity but produce limited expressive variations and weak subject specificity. 
Most existing approaches extract identity features from one or more reference images and inject them into a pretrained video generation backbone to guide subject-consistent generation. While this design effectively preserves appearance, the conditioning signal provides little or no information about the temporal dynamics that characterize how a particular individual naturally moves and expresses emotions. As a result, the generated facial behaviour is determined by the textual prompt and the generative prior of the pretrained model rather than by person-specific expressive patterns.

To address this challenge, we propose \ours, a reference-conditioned framework for person-specific video generation that jointly preserves visual identity and subject-specific expressive behaviour. Our approach complements the textual prompt and static identity reference image with a short reference video that serves as an explicit expressive conditioning signal. This design is motivated by the observation that both language and an image can describe the semantic content of an emotion, but often fail to specify how a particular individual expresses it through timing, intensity, gaze patterns, and head-motion dynamics. In contrast, a reference video naturally captures these rich spatio-temporal behavioural cues and provides direct information about the person's expressions.

By modeling identity and expressive behaviour as separate sources of variation, \ours enables independent control over appearance and behaviour during generation. It naturally encompasses a wider range of tasks, including identity preservation but also identity-expression transfer  as shown in \ref{fig:main}. It opens new opportunities for privacy-sensitive applications. In particular, it enables the generation of anonymized yet behaviourally faithful videos, as well as synthetic samples that can complement real-world datasets for guided training and evaluating human behaviour recognition systems.

The proposed framework builds upon the pretrained Wan video diffusion model \cite{wan2025wanopenadvancedlargescale} and augments it with complementary image and video conditioning branches, providing explicit identity and behavioural guidance during generation. Disentanglement between identity and expressive behaviour is primarily encouraged through the training data construction from the MEAD dataset \cite{kaisiyuan2020mead} and dedicated augmentation strategies, which expose the model with diverse combinations of identities and expressive behaviours. To integrate these conditioning signals, the image and video branches communicate with the generation pathway through an Identity-Expression Decoupled Attention (IEDA) mechanism combined with the Conditional Position Mapping (CPM) of StandIn \cite{xue2025stand}. This allows for controlled interactions between reference identity tokens, expressive tokens, and video tokens while preserving their distinct roles. With these methods, the model can effectively leverage complementary identity and behavioural signals while requiring only lightweight adaptation of the pretrained generator.


\noindent \textbf{Contributions:} (i) \ours -- a novel framework for expression-specific identity control in human-centric video generation. Unlike state-of-the-art identity-preserving approaches that primarily focus on appearance, our method explicitly focuses on expressivity by conditioning the generation on both a target identity and a reference expressive behaviour; (ii) By disentangling identity and expressive behaviour through dual image-video conditioning and cross-paired identity-expression training, our framework supports multiple inference-time tasks, including identity preservation and identity-expression transfer; (iii) Our experimental results obtained with the MEAD and RAVDESS datasets indicate that \ours can significantly improve the expressiveness of generated videos while maintaining state-of-the-art identity preservation.


\section{Related Work}

\noindent \textbf{(a) Video Generation:}
Current video generation models predominantly rely on diffusion-based frameworks, which have progressively shifted from U-Net architectures toward Diffusion Transformers (DiTs), as illustrated by large-scale Text-To-Video (T2V) foundation models such as Sora~\cite{lin2024opensoraplanopensourcelarge}, HunyuanVideo~\cite{kong2025hunyuanvideosystematicframeworklarge}, and Wan~\cite{wan2025wanopenadvancedlargescale}. These models synthesize realistic and temporally coherent videos from text descriptions and have demonstrated remarkable capabilities for general-purpose video generation. Although these models have demonstrated impressive capabilities for generating diverse, high-quality videos, maintaining both subject identity and fine-grained expression control remains a significant challenge.

\noindent \textbf{(b) Human-Centric Video Generation:} 
For finer control over the generated content, recent works \cite{lai2026slot, xue2025stand, liu2025phantomsubjectconsistentvideogeneration, yuan2025cons, jiang2025vaceallinonevideocreation} have moved from purely text-driven generation toward reference-conditioned paradigms. In Subject-to-Video (S2V), the model is given a text prompt and one or several reference images of the subject and aims to generate a video that both follows the textual instruction and preserves the visual attributes of the referenced subject.
Early S2V methods addressed identity preservation by jointly learning identity conditioning and video generation within a single end-to-end architecture. Representative examples include VACE~\cite{jiang2025vaceallinonevideocreation}, Phantom~\cite{liu2025phantomsubjectconsistentvideogeneration}, and ConsisID~\cite{yuan2025cons}. Although they differ in their implementation, these methods share a common principle: identity features are extracted from one or multiple reference images and injected directly into the generative backbone, allowing the model to learn subject-consistent video synthesis during training. This design has proven effective for preserving identity, establishing end-to-end identity conditioning as a strong baseline for S2V generation. However, end-to-end approaches require jointly learning both video generation and identity conditioning, which is computationally expensive. 

\noindent \textbf{(c) Lightweight Adapters for Identity Control:}
Rather than relearning the video generation capabilities of the backbone, recent works introduce lightweight identity adapters that augment frozen video foundation models, improving both training efficiency and modularity while facilitating adaptation to new architectures. ID-Animator~\cite{he2024idanimator} extracts identity embeddings from a reference image and injects them into a pretrained text-to-video model through a dedicated face adapter. To encourage the adapter to focus on identity rather than viewpoint, lighting, or background, it employs a random-reference training strategy using multiple images of the same individual. Stand-In~\cite{xue2025stand} introduces a lightweight plug-and-play identity adapter that preserves a person's identity from a single reference image while requiring only minimal additional training. Its modular design makes it particularly attractive for adapting recent video foundation models and serves as the basis of our work. Despite their efficiency and flexibility, these methods remain conditioned on static images. They preserve who the subject is but provide little information about how the subject naturally moves or expresses emotions.

To overcome this limitation, Slot-ID~\cite{lai2026slot} extends image-based identity adapters by conditioning video generation on both a reference portrait and a short reference video. The method extracts spatio-temporal features from the reference clip and compresses them into a compact set of identity tokens, which are subsequently integrated into a pretrained video generation backbone. By leveraging temporal information, Slot-ID achieves stronger identity consistency than single-image approaches while retaining the efficiency of adapter-based methods. However, while incorporating temporal references improves identity consistency, the temporal information is primarily used to build a richer identity representation. As a result, appearance and behavioural cues remain entangled, and the expressive dynamics contained in the reference video are not explicitly modeled as a separate controllable factor. This limitation highlights a broader gap in current S2V methods. Most existing approaches are designed to preserve who the subject is, rather than how the subject behaves. Although recent S2V models have made substantial progress in maintaining identity consistency, they remain limited in their ability to faithfully preserve person-specific expression dynamics.

Instead of simply increasing the amount of reference information available to the model, this paper advocates for explicit disentanglement of visual identity from expressive behaviour, so that the generation model can preserve either factor independently or both jointly.
\section{Proposed \ours Method}

\begin{figure}[!b]
    \centering
    \includegraphics[width=1\linewidth]{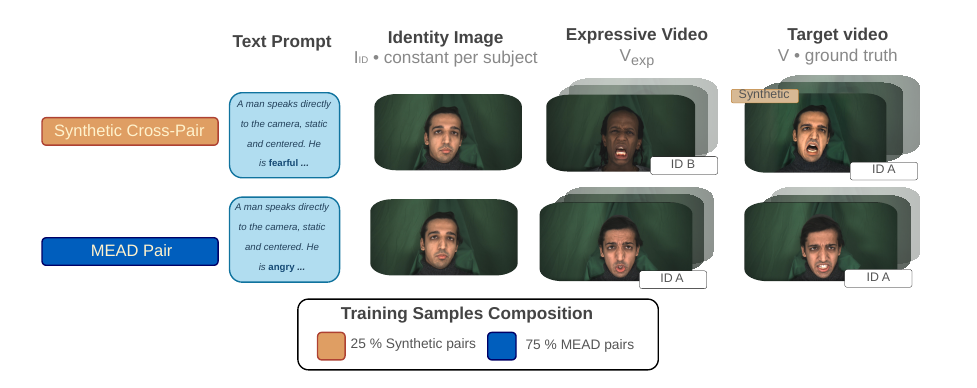}
    \caption{\textbf{\ours training scheme.} Training combines real MEAD pairs, where the identity image, expressive video, and target video all belong to the same subject, with synthetic cross-identity pairs, where the expressive video contains a different identity and the target video is generated using a Portrait Animation Model~\cite{xu2025hunyuanportrait}.}
    \label{fig:training}
\end{figure}

We address S2V generation by explicitly factorizing the subject conditioning $S$ into identity and expressive behaviour such that the generation process becomes:
\begin{equation}
p_\theta(V \mid p, S)
\quad\longrightarrow\quad
p_\theta(V \mid p, I_{\mathrm{id}}, V_{\mathrm{exp}}),
\label{eq:conditioning}
\end{equation}
where $p$ denotes the text prompt, $I_{\mathrm{id}}$ is a reference image encoding visual identity and $V_{\mathrm{exp}}$ is a reference video encoding expressive behaviour. This formulation enables independent control over appearance and expressive dynamics. \ours implements this formulation by introducing two complementary components: an expressive video conditioning branch (Sec.~\ref{sec:3_expressive_video_conditioning_branch}, Fig.~\ref{fig:architecture}) and a dedicated training scheme (Sec.~\ref{sec:3_expressive_video_conditioning_branch}, Fig.~\ref{fig:training}) designed to promote the disentanglement of visual identity and expressive behaviour that together enable the generation of more expressive videos while independently controlling visual identity and expressive behaviour.

\subsection{Background}
\label{sec:3_background}

\noindent\textbf{Stand-In}~\cite{xue2025stand} is a lightweight S2V framework that augments pretrained flow-based video generators with an identity conditioning branch. Rather than fine-tuning the entire backbone, it introduces a plug-and-play identity adapter that injects information from a single reference image into the latent generation process. This modular design achieves strong identity preservation while requiring only minimal adaptation of the pretrained model, making it a particularly suitable foundation for our proposed identity-expression factorization.

The model reuses the pretrained VAE of the video backbone to encode the reference image into the same latent space as the generated video. Given a reference image $I_{\text{id}}$, Stand-In computes image latents $z_r = E_{\mathrm{VAE}}(r)$, then patchifies them into image tokens $T_{\text{id}}$. In parallel, the noisy video latents are also patchified into video tokens $T_v$. The two token sequences $T_{\text{id}}$ and $T_v$ are concatenated and processed together inside the DiT blocks.

Identity information is transferred from the reference image to the generated video through \emph{Restricted Self-Attention}. The key idea is to allow video tokens to attend to reference-image tokens, while preventing the reference tokens from being updated by the noisy video tokens. This keeps the reference branch stable and avoids treating the conditioning image as part of the denoising video sequence. Stand-In also introduces \emph{CPM}, which assigns the reference-image tokens their own positional coordinate system. This avoids forcing image and video tokens to share the same spatio-temporal positions, which would disturb the pretrained positional prior of the video backbone.

\subsection{Expressive Video Conditioning Branch}
\label{sec:3_expressive_video_conditioning_branch}

\begin{figure*}[!b]
    \centering
    \includegraphics[width=\textwidth]{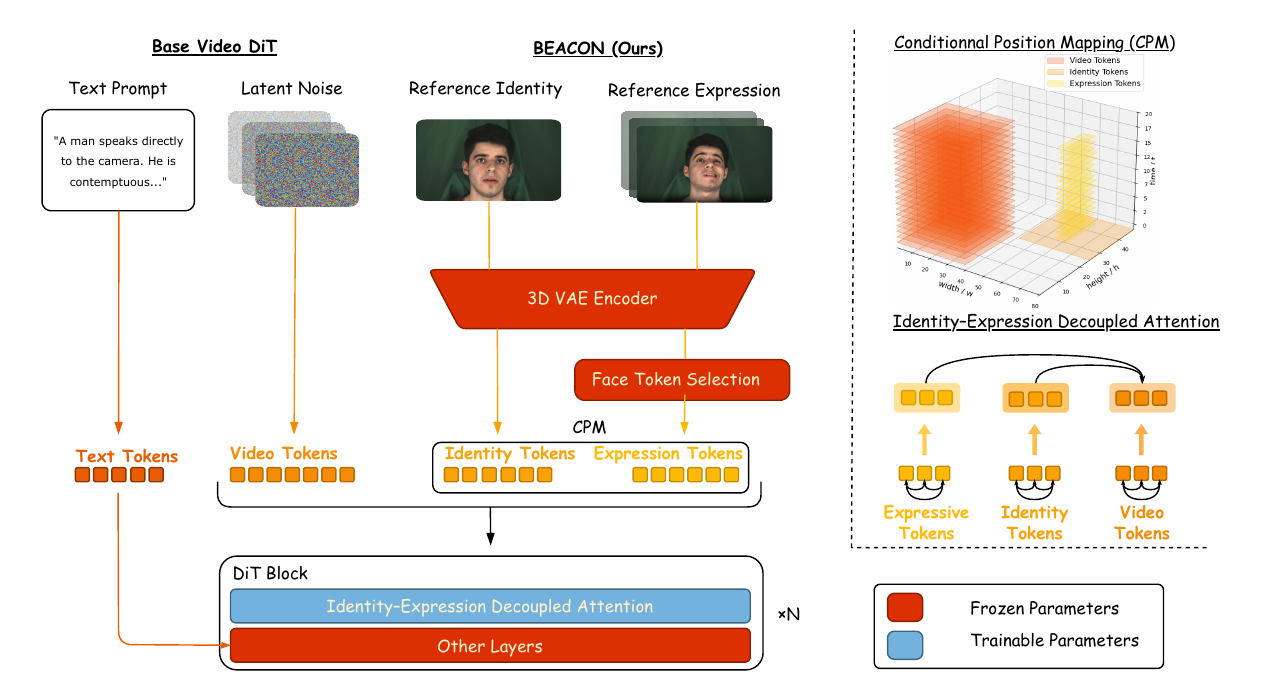}
    \caption{Overview of the proposed \ours architecture. It conditions T2V generation on an identity reference image and an expressive reference video in addition to the text prompt. The expressive branch extracts facial dynamics from the reference video and injects them into the frozen video generation backbonwhile preserving their
    decoupling from visual identity.
    }
    \label{fig:architecture}
\end{figure*}

We introduce a secondary conditioning stream based on an expressive reference video. While the identity branch captures static appearance cues, the expressive branch provides dynamic information such as facial motion, expression intensity, head movement, and the temporal evolution of emotions. The noisy latent video $z_t$ is patchified into video tokens $T_v$. In parallel, the identity reference image and the expressive reference video are encoded using the pretrained VAE of the video backbone. The model therefore operates on three token groups: video tokens $T_v$, identity tokens $T_{\mathrm{id}}$, and expressive tokens $T_{\mathrm{exp}}$, corresponding respectively to generation, identity, and expressive behaviour. Both conditioning branches are added on top of the pretrained T2V backbone, while keeping the generative backbone frozen.

\noindent \textbf{Facial Token Selection.}
The reference video produces a large number of spatio-temporal latent tokens. Since our objective is to model facial expressive behaviour rather than scene dynamics, many of the video tokens are irrelevant. We use InsightFace \cite{deng2019arcface} to localize the face area and retain only the expressive tokens associated with facial regions. Let $\mathcal{F}$ denote the set of token locations belonging to the detected face region. The filtered expressive token set is defined as:
\begin{equation}
T_{\mathrm{exp}}^{f}
=
\left\{
t_i \in T_{\mathrm{exp}}
\;\middle|\;
\mathrm{loc}(t_i)\in\mathcal{F}
\right\}.
\label{eq:facial_tokens}
\end{equation}

This filtering step reduces the number of expressive tokens from $16.3$k to $2.2$k while preserving the most relevant expression-related information. The resulting token set $T_{\mathrm{exp}}^{f}$ is subsequently used as the expressive conditioning signal.

\noindent \textbf{Identity-Expression Decoupled Attention.}
Our proposed attention mechanism operates on three token groups: generated video tokens $T_v$, identity tokens $T_{\mathrm{id}}$, and expressive tokens $T_{\mathrm{exp}}^{f}$, as illustrated in Fig.~\ref{fig:architecture}. To this end, generated video tokens are allowed to attend to both conditioning branches, enabling the synthesis process to integrate appearance and expressive information simultaneously. In contrast, the identity and expressive branches remain completely isolated and never attend to one another, preventing information leakage between the two conditioning signals. Video tokens attend to all available token groups,
\begin{equation}
T_v
\leftarrow
\mathrm{Attn}
\Big(
Q_v,
[K_v, K_{\mathrm{id}}, K_{\mathrm{exp}}],
[V_v, V_{\mathrm{id}}, V_{\mathrm{exp}}]
\Big)
\label{eq:decoupled_attention}
\end{equation}
while each conditioning branch remains isolated and performs self-attention independently,
\begin{equation}
\begin{aligned}
T_{\mathrm{id}}
&\leftarrow
\mathrm{Attn}
(Q_{\mathrm{id}}, K_{\mathrm{id}}, V_{\mathrm{id}}), \\
T_{\mathrm{exp}}^{f}
&\leftarrow
\mathrm{Attn}
(Q_{\mathrm{exp}}, K_{\mathrm{exp}}, V_{\mathrm{exp}}).
\end{aligned}
\label{eq:identity_expression_attention}
\end{equation}

\noindent \textbf{Extending Conditional Position Mapping.}
We extend CPM (introduced in Sec.~\ref{sec:3_background}) to support the additional expressive conditioning branch while preserving the pretrained spatio-temporal positional prior of the video backbone. Generated video tokens retain their original coordinates, whereas both conditioning branches are projected into a dedicated region. Let $(t,x,y)$ denote the temporal and spatial coordinates of a video token. Generated video tokens retain their original coordinates:
\begin{equation}
P_v = (t, x, y).
\label{eq:video_position}
\end{equation}

The identity and expressive conditioning branches are projected into a dedicated conditioning region of the positional space through a common spatial offset $(h,w)$. Identity tokens correspond to a static image and therefore receive a fixed artificial temporal coordinate:
\begin{equation}
P_{\mathrm{id}}
=
(-1,\;x+h,\;y+w)
\label{eq:identity_position}.
\end{equation}

In contrast, expressive tokens preserve their temporal structure:
\begin{equation}
P_{\mathrm{exp}}
=
(t,\;x+h,\;y+w).
\label{eq:expression_position}
\end{equation}

As a result, both conditioning branches occupy the same spatial region while remaining distinguishable through their temporal coordinates. This design allows the model to preserve the temporal structure of the expressive reference video while clearly separating conditioning tokens from generated video tokens.

\subsection{Training Scheme}
\label{sec:3_training_scheme}

A key contribution of \ours lies in the construction of the training pairs, which are specifically designed to encourage the disentanglement of visual identity and expressive behaviour without requiring any explicit supervision. Rather than pairing the target video with an arbitrary reference sequence, we exploit the structure of the MEAD dataset, where each subject performs multiple recordings of the same emotion at different intensity levels.

For every target video $V$, the identity branch always receives the same neutral reference image $I_{\mathrm{id}}$ of the corresponding subject. The identity pathway is consistently exposed to stable appearance cues throughout training. The expressive branch is conditioned on a different recording $V_{\mathrm{exp}}$ of the same subject performing the same emotion with the same intensity as the target video. Although both videos share the same expressive behaviour, they correspond to different performances, preventing the model from trivially copying the target sequence while still providing meaningful supervision of subject-specific facial dynamics.

This training protocol naturally encourages the desired factorization. Because the identity branch always observes the same neutral portrait, identity information is associated with the appearance pathway. In contrast, the expressive branch receives temporally varying recordings of the same subject, encouraging it to encode behavioural patterns rather than static visual characteristics.

\noindent \textbf{Synthetic Cross-Pair Regularization.}
To further reinforce the disentanglement between appearance and behaviour, the training set is augmented with synthetic expressive videos generated using HunyuanPortrait~\cite{xu2025hunyuanportrait}. For these synthetic samples, the driving motion is taken from a different identity than that of the reference image, ensuring that the identity encoded in the expressive video intentionally differs from the target identity. This prevents the expressive branch from recovering appearance cues, forcing it to encode only behavioural dynamics. Synthetic samples account for 25\% of each training mini-batch, while the remaining 75\% are real MEAD videos. This regularization encourages the model to consistently recover the subject's appearance from the identity image while relying on the expressive branch exclusively for facial motion and expression.
\section{Experimental Methodology}

\subsection{Implementation Details}

\ours is built upon a 14B DiT T2V-WanVideo model~\cite{wan2025wanopenadvancedlargescale}. We fine-tune the model using rank-128 LoRA applied to the QKV projection of the conditioning tokens in the self-attention layer, introducing around 0.15B trainable parameters. The training is performed using the AdamW optimizer with a learning rate of $1 \times 10^{-4}$. We train the model for 3k steps. All videos contain 81 frames. Training is performed on approximately 2,000 identity-expression pairs extracted from the MEAD dataset, including 500 synthetic pairs generated with mismatched identity and expressive behaviour for cross-pair regularization. Videos are resized to $640\times384$. The model is trained with a per-GPU batch size of 1 on four NVIDIA A100 GPUs and a gradient accumulation factor of 2, resulting in an effective batch size of 8. Additionally, the expressive conditioning branch is randomly dropped with a probability of 10\% during training. Empirically, this regularization mitigates high-frequency artifacts introduced by the expressive reference video and results in more stable and visually coherent generations.

\subsection{Dataset}

We evaluate \ours on the MEAD dataset~\cite{kaisiyuan2020mead}, a controlled facial expression dataset containing multiple identities performing the same scripted sentences under eight emotion categories and three intensity levels. Its controlled acquisition protocol provides repeated performances of the same emotion by each subject, making it particularly well suited for evaluating both identity preservation and expressive behaviour.

In addition to MEAD, we evaluate \ours on the RAVDESS dataset~\cite{livingstone2018ravdess} to assess its generalization to a different facial expression dataset. RAVDESS contains recordings of 24 professional actors (12 female and 12 male) performing eight emotional expressions: neutral, calm, happy, sad, angry, fearful, disgust, and surprised, with two intensity levels for most emotions. Its different subjects and expressive patterns provide a complementary setting for evaluating the generalization of our method beyond MEAD.

\subsection{Evaluation Metrics}

\ours is evaluated according to three complementary dimensions: identity preservation, expressive behaviour consistency, and video quality.

\noindent \textbf{Identity Preservation.} Identity preservation is measured using the Face Similarity (FaceSim) metric from the OpenS2V benchmark~\cite{yuan2025opensv}. This benchmark follows the protocol first introduced by ConsisID~\cite{yuan2025cons}: InsightFace is first used to detect and align faces in both the generated frames and the reference image. Identity embeddings are then extracted using CurricularFace, and cosine similarities are averaged over all valid frames to obtain a single FaceSim score for each generated video.

\noindent \textbf{Expressive Behaviour Consistency.}
To quantify how faithfully facial dynamics are reproduced, we use the Expressive Fréchet Inception Distance (E-FID), recently adopted for audio-driven portrait synthesis~\cite{tian2024emo,chen2024echomimic}. Following~\cite{deng2020deep3d}, facial expression parameters are first extracted from each frame using a 3D face reconstruction model. E-FID is then computed as the Fréchet distance between the distributions of expression parameters extracted from the generated and ground-truth videos. Lower values indicate better agreement between the generated and target expressive behaviours. We additionally report the Action Unit Error (AUE), a metric used to evaluate facial motion accuracy in talking-head generation~\cite{talkinggaussian,dgtalker}. Action Unit intensities are extracted using OpenFace~\cite{baltrusaitis2018openface} and AUE is computed as the average absolute difference between generated and ground-truth AU vectors.

\noindent \textbf{Video Quality.}
We assess video quality using VBench~\cite{huang2024vbench}, reporting Subject Consistency, Aesthetic Quality, and Motion Smoothness to evaluate appearance consistency, perceptual quality, and temporal coherence, respectively.

\noindent \textbf{Prompt.}
All methods are evaluated using the same prompt template. It specifies the target emotion while keeping the scene configuration unchanged, ensuring that differences in performance arise from the conditioning strategy rather than prompt variations. For example, the prompt used for an angry sequence is:
\begin{quote}
``A man speaks directly to the camera. The camera is static and centered. He is angry, with a strong facial expression and high emotional intensity. Lighting is even and frontal, with a clean and simple background.''
\end{quote}

\noindent \textbf{Choice of Reference Image.}
For each subject, we select a neutral frontal frame as identity reference to minimize expression, pose, viewpoint, and lighting-specific cues.

\subsection{Evaluation Protocol}

Generating a single S2V video requires approximately 6 to 10 minutes on a single A100 GPU. Therefore, exhaustive evaluation over thousands of videos would be prohibitively expensive. To obtain reliable estimates for distribution-based metrics such as E-FID while keeping the computational cost reasonable, we evaluate all methods on a selected subset of the MEAD dataset. We evaluate eight held-out subjects (four female, four male), ensuring evaluation on unseen identities. For each subject, we generate ten videos for each of the eight emotions, resulting in 640 videos (80 per subject and emotion).

\section{Results and Interpretation}

\subsection{Comparison with State-of-the-art Methods}

Table~\ref{tab:quant_comparison} reports the comparison between \ours and existing methods on the MEAD dataset. \ours substantially outperforms other methods in terms of expressive behaviour consistency. This confirms that conditioning on a reference video, rather than a single static image, provides the model with the temporal cues necessary to preserve person-specific expressive dynamics.

In terms of identity preservation, \ours achieves a FaceSim of 79.0\%, slightly below Stand-In's 82.6\%. However, this comparison relies on a neutral reference frame of the subject. As shown in Table~\ref{tab:facesim}, when FaceSim is instead computed against a reference frame sampled from the same emotion as the generated video, the trend reverses: \ours's FaceSim increases while Stand-In's drops substantially. This discrepancy suggests that Stand-In tends to generate videos with limited expressive variation, remaining close to a neutral appearance regardless of the target emotion, whereas \ours produces expressive videos that more faithfully reflect the intended emotional state.

\begin{table*}[h]
\centering
\caption{Quantitative comparison of S2V generation methods on the MEAD dataset.}
\label{tab:quant_comparison}

\fontsize{6.5}{7.5}\selectfont
\setlength{\tabcolsep}{1.5 pt}
\renewcommand{\arraystretch}{1.1}

\begin{tabular}{@{}lccccccc@{}}
\toprule
\multirow{2}{*}{\textbf{Method}} &
\multicolumn{2}{c}{\textbf{E-FID} $\downarrow$} &
\multirow{2}{*}{\textbf{AUE} $\downarrow$} &
\multirow{2}{*}{\textbf{FaceSim} $\uparrow$} &
\multirow{2}{*}{\shortstack{\textbf{Subject}\\\textbf{Consistency} $\uparrow$}} &
\multirow{2}{*}{\shortstack{\textbf{Aesthetic}\\\textbf{Quality} $\uparrow$}} &
\multirow{2}{*}{\shortstack{\textbf{Motion}\\\textbf{Smoothness} $\uparrow$}} \\
\cmidrule(lr){2-3}
&
\textbf{Per Subject} &
\textbf{Per Emotion} \\
\midrule

Stand-In~\cite{xue2025stand}
& 2.28
& 2.87
& 0.259
& \textbf{0.826}
& 0.985
& \textbf{0.526}
& 0.996 \\

ConsisID~\cite{yuan2025cons}
& 3.17
& 3.94
& 0.295
& 0.804
& 0.948
& 0.502
& 0.982 \\

BEACON
& \textbf{0.86}
& \textbf{0.89}
& \textbf{0.142}
& 0.760
& \textbf{0.991}
& 0.500
& \textbf{0.997} \\

\bottomrule
\end{tabular}
\end{table*}
\begin{table}[h]
\centering
\small
\setlength{\tabcolsep}{6pt}
\renewcommand{\arraystretch}{1.1}

\caption{Face similarity evaluated with neutral and expressive reference frames.}
\label{tab:facesim}

\begin{tabular}{llc}
\toprule
\textbf{Method} &
\textbf{Ground Truth Frame} &
\textbf{FaceSim $\uparrow$} \\
\midrule

Stand-In~\cite{xue2025stand}
& Neutral
& \textbf{0.826}\\

BEACON
& Neutral
& 0.760 \\

\midrule

Stand-In~\cite{xue2025stand}
& Expressive
& 0.678  \\

BEACON
& Expressive
& \textbf{0.852}\\

\bottomrule
\end{tabular}
\end{table}

\begin{table*}[!t]
\centering
\caption{Comparison of S2V generation methods on the RAVDESS dataset.}
\label{tab:quant_comparison_ravdess}

\fontsize{6.5}{7.5}\selectfont
\setlength{\tabcolsep}{1.5 pt}
\renewcommand{\arraystretch}{1.1}

\begin{tabular}{@{}lccccccc@{}}
\toprule
\multirow{2}{*}{\textbf{Method}} &
\multicolumn{2}{c}{\textbf{E-FID} $\downarrow$} &
\multirow{2}{*}{\textbf{AUE} $\downarrow$} &
\multirow{2}{*}{\textbf{FaceSim} $\uparrow$} &
\multirow{2}{*}{\shortstack{\textbf{Subject}\\\textbf{Consistency} $\uparrow$}} &
\multirow{2}{*}{\shortstack{\textbf{Aesthetic}\\\textbf{Quality} $\uparrow$}} &
\multirow{2}{*}{\shortstack{\textbf{Motion}\\\textbf{Smoothness} $\uparrow$}} \\
\cmidrule(lr){2-3}
&
\textbf{Per Subject} &
\textbf{Per Emotion} \\
\midrule

Stand-In~\cite{xue2025stand}
& 2.89
& 1.83
& 0.202
& 0.805
& 0.990
& \textbf{0.527}
& \textbf{0.996} \\

ConsisID~\cite{yuan2025cons}
& 3.97
& 2.85
& 0.248
& 0.802
& 0.958
& 0.523
& 0.991 \\

BEACON
& \textbf{1.57}
& \textbf{0.97}
& \textbf{0.174}
& \textbf{0.854}
& \textbf{0.993}
& 0.522
& \textbf{0.996} \\

\bottomrule
\end{tabular}
\end{table*}

\ours generalizes well to other talking-head datasets, as illustrated by the results on RAVDESS in Table~\ref{tab:quant_comparison_ravdess}, where \ours still outperforms Stand-In on expressivity. Interestingly, the FaceSim trend is reversed on RAVDESS. This may be explained by the high sensitivity of face recognition models to acquisition conditions such as pose and illumination.

\begin{figure}[!t]
    \centering
    \includegraphics[width=1\linewidth]{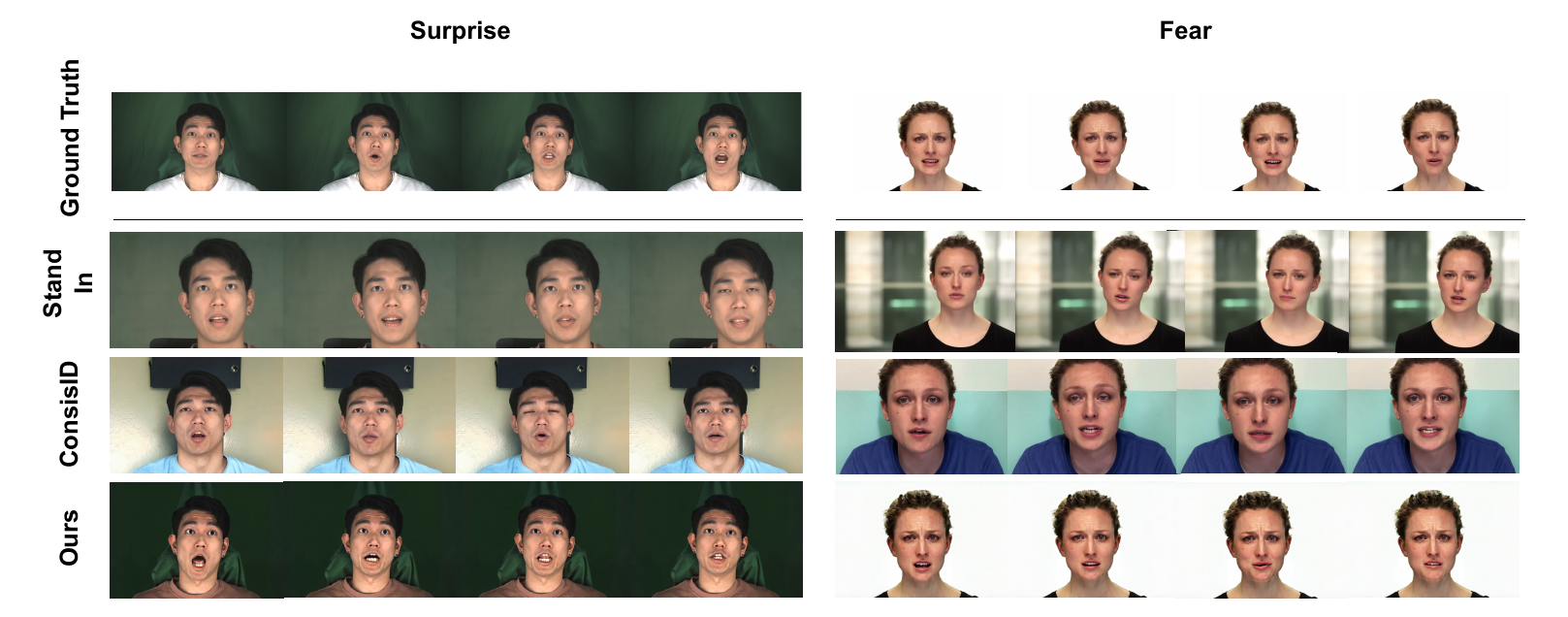}
    \caption{\textbf{Qualitative comparison of generated videos.} Examples are shown from two datasets: MEAD (left) and RAVDESS (right).}
    \label{fig:qual_results}
\end{figure}

\begin{figure}[!t]
    \centering
    \includegraphics[width=1\linewidth]{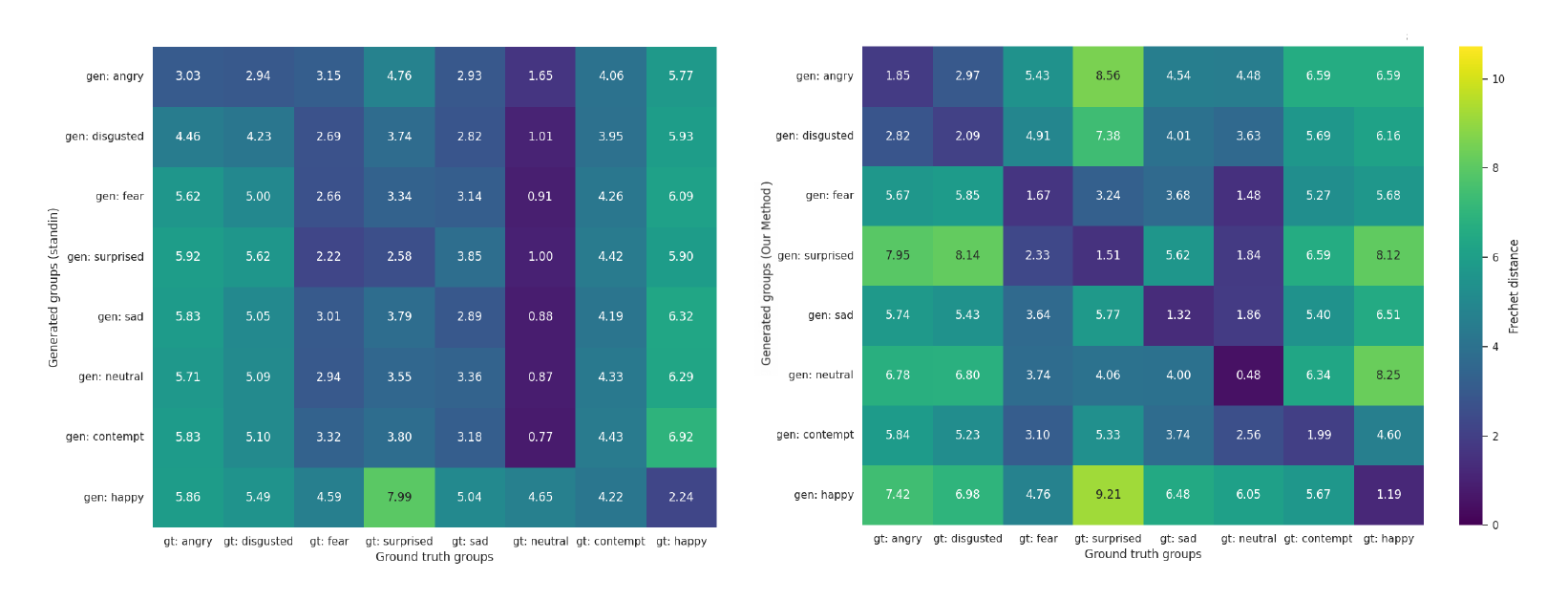}
    \caption{\textbf{E-FID matrices} between generated videos and ground-truth videos across all emotion categories on MEAD dataset. \textbf{Left:} Stand-In. \textbf{Right:} \ours.}
    \label{fig:matrix}
\end{figure}

Qualitative comparisons are presented in Fig.~\ref{fig:qual_results}. Both Stand-In and ConsisID tend to generate behaviourally neutral videos, whereas \ours produces richer and more subject-specific facial deformations. In the \textit{fear} generation example, Stand-In fails to reproduce some of the subject's characteristic facial movements, such as the forehead wrinkles visible in the expressive reference. In contrast, \ours better captures these fine-grained facial dynamics by leveraging the expressive video conditioning. 
This tendency to generate behaviourally neutral videos is also reflected in the E-FID matrices in Fig.~\ref{fig:matrix}, which compare the generated videos with reference videos of the same subjects across different emotions. For Stand-In, regardless of the target emotion, the generated videos consistently exhibit a lower E-FID with the subject's neutral videos than with videos of the corresponding emotion, indicating a bias toward neutral facial behaviour. In contrast, \ours achieves the lowest E-FID for the correct emotion, demonstrating its ability to reproduce subject-specific expressive behaviour rather than collapsing toward neutral expressions.

\subsection{Analysis of Disentanglement}

Table~\ref{tab:synthetic_ablation} reports results on 80 videos generated from the MEAD evaluation set, where the identity image and expressive video are taken from different subjects to evaluate identity-expression transfer. This experiment assesses whether the model extracts expressive dynamics from the expressive video while preserving the visual identity provided by the identity image. The results demonstrate that synthetic cross-identity training pairs are essential for learning effective identity-expression disentanglement. They increase the Face Similarity between the generated video and the identity reference from 0.10 to 0.69, while reducing the similarity with the expressive reference identity from 0.67 to 0.25, indicating substantially less identity leakage from the expressive branch. Without synthetic cross-identity training, the model instead generates videos that resemble the identity of the expressive reference rather than that of the identity reference.

\begin{table}[!b]
\centering
\small
\setlength{\tabcolsep}{4pt}
\renewcommand{\arraystretch}{1.1}

\caption{Ablation of synthetic training data for identity-expression disentanglement}
\label{tab:synthetic_ablation}

\begin{tabular}{lccc}
\toprule
\textbf{Method} &
\multicolumn{2}{c}{\textbf{Face Similarity}} \\
\cmidrule(lr){2-3}
&
\textbf{Identity Ref.} $\uparrow$ &
\textbf{Expression Ref.} $\downarrow$ &
\textbf{E-FID} $\downarrow$ \\
\midrule

BEACON w/o Synthetic Data & 0.10 & 0.67 & \textbf{0.83}\\

BEACON & \textbf{0.69} & \textbf{0.25} & 1.01 \\

\bottomrule
\end{tabular}
\end{table}

The slight increase in E-FID can be explained by the role of synthetic cross-identity training. Without synthetic cross-identity pairs, the model mainly learns to reproduce the expressive reference video, effectively copying both its identity and facial dynamics. This behavior is also reflected in the Face Similarity metrics, where the generated videos remain substantially more similar to the expressive reference identity. In contrast, introducing synthetic cross-identity data forces the model to transfer expressive behaviour to a different identity rather than directly replicating the reference video. This more challenging objective may introduce slight deviations in the transferred facial dynamics, leading to a marginal increase in E-FID. 

\begin{figure}[!t]
    \centering
    \includegraphics[width=1\linewidth]{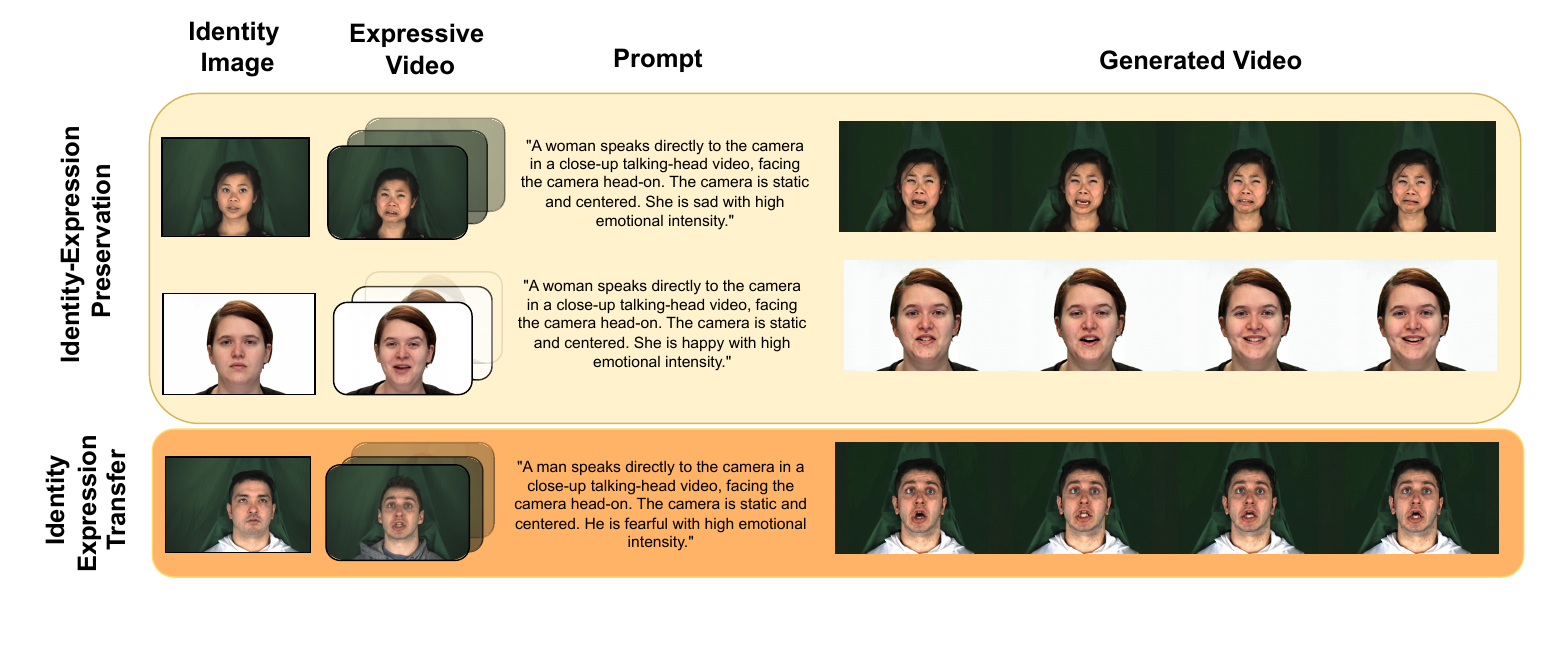}
    \caption{\textbf{Identity-expression preservation and transfer.} Examples of the different inference settings enabled by \ours. Identity and expressive references can originate from the same subject for joint preservation, or from different subjects for identity-expression transfer.}
    \label{fig:examples}
\end{figure}

Although the model successfully transfers expressive behaviour across identities, as illustrated in Fig.~\ref{fig:main}, minor appearance leakage may still occur in some cases, as shown in Fig.~\ref{fig:examples}, where the generated eye color is partially influenced by the expressive reference. This residual leakage is also reflected in the non-zero Face Similarity with the expressive reference identity.
This disentanglement also enables different inference settings, as illustrated in Fig.~\ref{fig:examples}. When the identity image and expressive video belong to the same subject, the model jointly preserves identity and expressive behaviour. Otherwise, when the two conditioning signals come from different subjects, the generated video combines the identity of the reference image with the expressive behaviour of the reference video, resulting in identity-expression transfer.
\section{Conclusion}

\ours is a lightweight framework proposed for person-specific video generation that jointly conditions on an identity image and an expressive reference video. By separating identity and expressive behaviour through dual conditioning and a dedicated training strategy, \ours generates videos with more subject-specific facial dynamics while maintaining strong identity preservation. Experiments on MEAD and RAVDESS display improved expressive behaviour consistency compared with existing S2V approaches, while the proposed factorization also supports identity-expression transfer. Our framework focuses on controlled talking-head videos with static cameras and relatively simple backgrounds. In contrast, one of the strengths of recent S2V models is their ability to generate subjects across diverse scenes and poses. Extending \ours beyond controlled talking-head settings to more diverse and unconstrained scenarios therefore represents an important direction for future work.

%
%
\clearpage
\bibliographystyle{splncs04}
\bibliography{main}

\end{document}